\documentclass[conference]{IEEEtran}
\IEEEoverridecommandlockouts
\usepackage{cite}
\usepackage{amsmath,amssymb,amsfonts}
\usepackage{algorithmic}
\usepackage{graphicx}
\usepackage{textcomp}
\usepackage{xcolor}
\def\BibTeX{{\rm B\kern-.05em{\sc i\kern-.025em b}\kern-.08em
    T\kern-.1667em\lower.7ex\hbox{E}\kern-.125emX}}
\begin{document}

\title{Zero-Shot Cross-Lingual Transfer in Legal Domain using Transformer Models}

\author{\IEEEauthorblockN{1\textsuperscript{st} Zein Shaheen}
\IEEEauthorblockA{
\textit{ITMO University}\\
Russia, Saint Petersburg \\
shaheen@itmo.ru}
\and
\IEEEauthorblockN{2\textsuperscript{nd} Gerhard Wohlgenannt}
\IEEEauthorblockA{
\textit{ITMO University}\\
Russia, Saint Petersburg \\
gwohlg@corp.ifmo.ru}
\and
\IEEEauthorblockN{3\textsuperscript{rd} Dmitry Mouromtsev}
\IEEEauthorblockA{
\textit{ITMO University}\\
Russia, Saint Petersburg \\
mouromtsev@itmo.ru}

}

\maketitle

\begin{abstract}
Zero-shot cross-lingual transfer is an important feature in modern NLP models and architectures to support low-resource languages.
In this work, We study zero-shot cross-lingual transfer from English to French and German under Multi-Label Text Classification, where we train a classifier using English training set, and we test using French and German test sets. We extend EURLEX57K dataset, the English dataset for topic classification of legal documents, with French and German official translation. We investigate the effect of using some training techniques, namely Gradual Unfreezing and Language Model finetuning, on the quality of zero-shot cross-lingual transfer. We find that Language model finetuning of multi-lingual pre-trained model (M-DistilBERT, M-BERT) leads to 32.0-34.94\%, 76.15-87.54\% relative improvement on French and German test sets correspondingly. Also, Gradual unfreezing of pre-trained model's layers during training results in relative improvement of 38-45\% for French and 58-70\% for German. Compared to training a model in Joint Training scheme using English, French and German training sets, zero-shot BERT-based classification model reaches 86\% of the performance achieved by jointly-trained BERT-based classification model. 
\end{abstract}


\section{Introduction}
Cross-lingual transfer learning provides a way to train a model using a dataset in one or more languages and use this model to make inferences in other languages. This type of transfer learning can benefit applications such as question answering \cite{lee2019cross}, dialogue systems  \cite{schuster2018cross}, machine translation \cite{ji2020cross}, named entity recognition \cite{johnson2019cross}, as in all of these applications it is essential to have good representations of words and texts. These representations should be independent of the language and capture high-level semantic relations.

Contextual word embeddings (such as ELMo \cite{peters2018deep}, GPT \cite{radford2018improving}, or BERT \cite{devlin2018bert}) have shown state-of-the-art performance on many NLP tasks. Their performance depends on the availability of a large amount of labeled text data. Recent work with Multilingual BERT (M-BERT) has demonstrated that the model performs well in zero-shot settings \cite{conneau2018xnli}. In this case, only labeled English data are necessary to train the model and use it to make inferences in another language.

Large-scale Multi-label Text Classification (LMTC) is the task of assigning a subset from a collection of thousands of labels to a given document. There are many challenges connected with this task. First, the distribution of labels is usually sparse and follows the power-law distribution. Another challenge is the availability of a large dataset to train a good model that generalizes well to unseen data. Collecting and annotating such datasets is an expensive and cumbersome process; annotators need to read the entire document and check against all available labels to decide which labels to assign to the document. Furthermore, it is very likely that annotators are missing some potentially correct tags.

Cross-lingual transfer learning (CLTL) can mitigate the issue of dataset availability for LMTC tasks by jointly training an LTMC model for several languages. It is also possible to train an LTMC for low-resources languages in zero-shot settings using available data in other languages and then making inferences in the unseen target language.

French and German alongside with English are the main focus of this paper. Ethnologue's method of calculating lexical similarity between languages \cite{rensch1992calculating} shows that English has a lexical similarity of 60\% with German and 27\% with French. Ethnologue's method compares a regionally standardized wordlist and counts those forms that show similarity in both form and meaning. 

In this work, we focus on cross-lingual transfer learning for LMTC task, based on JRC-Acquis dataset \cite{steinberger2006jrc} and an extended version of EURLEX57K \cite{chalkidis2019large} dataset. Both datasets contain documents from EurLex, the legal database of the European Enion (EU), and they are annotated using descriptors from the the European Union’s multilingual and multidisciplinary thesaurus EuroVoc. JRC-Acquis is a large parallel corpus of documents available in 25 languages including English, French and German. EURLEX57K is available in English, we extended this dataset to include parallel documents in French and German.

The goal of this work is to start a baseline for LMTC based on these two multilingual datasets which contain parallel documents in English, French and German. We compare between two CLTL settings for this task: (i) a zero-shot setting in which we train a multi-lingual model using the English training set and then we test using the French and German test sets; (ii) a joint training setting in which we train the model using all training data including English, French and German training sets.

The main findings and contributions of this work are: (i) the experiments with multilingual-BERT and multilingual-DistilBERT with gradual unfreezing and language model finetuning (ii) providing a new standardized multilingual dataset for further investigation, (iii) ablation studies to measure the impact and benefits of various training strategies.

The remainder of the paper is organized as follows: After a discussion of related work in Section \ref{sec-relatedworks}, we discuss CLTL (section \ref{sec-cross-lingual}) and multi-lingual datasets (sections \ref{sec-datasets}). Then we present the main methods (BERT, DistilBERT) and strategies for training multi-lingual model in Section \ref{section-methods}. Section \ref{sec_results} contains extensive evaluations of the methods on both datasets as well as ablation studies, and after a discussion of results (Section \ref{sec-discussion}) we conclude the paper in Section \ref{sec-conclusion}.

\section{Related Works}
\label{sec-relatedworks}
In the realm of cross-lingual transfer learning, Eriguchi et al.~\cite{eriguchi2018zero} performed zero-shot binary sentiment classification by reusing an encoder from multilingual neural machine translation; they extended this encoder with a task-specific classifier component to perform text classification in a new language, where training data in this particular language was not used. On Amazon reviews, their model achieves 73.88\% accuracy on the French test set in zero-shot settings when training using English training data only, meanwhile including French training data in the training process increases the accuracy on the French test set to 83.10\%. As a result, the zero-shot model obtains 92.8 \% of the accuracy achieved after including French training data. \\
Pelicon et al.~\cite{pelicon2020zero} used multilingual BERT to perform zero-shot sentiment classification by training a classifier in Slovene and making inference using texts in other languages. The model trained using the Slovene training set obtains $52.41 \pm2.58$ F1-score on the Croatian test set, however on the Slovene test set its performance reaches $63.39 \pm 2.42$ F1-score.\\
Keung et al.~\cite{keung2019adversarial} improved zero-shot cross-lingual transfer learning for text classification and named entity recognition by incorporating language-adversarial training to extract language-independent representations of the texts and align the embeddings of English documents and their translations. Regarding the classification task, they trained a classifier using English training data of the MLDoc dataset, they report  85.7\% and 88.1\% accuracy on French and German test sets correspondingly after using language-adversarial training.\\
Chalkidis et al.~\cite{chalkidis2019large} published a new EURLEX57K dataset, a dataset of European legal documents in English. Steinberger et al.~\cite{steinberger2006jrc} presented JRC-Acquis, a freely available parallel corpus containing European Union documents. This dataset is available in 20 official EU languages, including English, French, and German.\\
In our previous work~\cite{shaheen2020large}, we used a transformer-based pre-trained model (BERT, DistilBERT, RoBerta, XLNet) to extract high-level vector representations from legal documents. First, we applied Language Model Finetuning (LMFT)  to this model using documents from the training set; the goal here is to improve the quality of document representations extracted from this model. Then, we extened the previously finetuned model with a classifier. Later, the transformer model and the classifier were jointly trained while gradually unfreezing the layers of the transformer model during training. This approach led to a significant improvement in the quality of the model.

In this work, we experiment with Multilingual-BERT and Multilingual-DistilBERT under cross-lingual zero-shot and joint-training transfer settings. We provide ablation studies to measure the impact of various training strategies and heuristics. Moreover, we provide new standardized multilingual dataset for further investigation by the research community.

\section{Cross-Lingual Transfer Learning}
\label{sec-cross-lingual}
The idea behind Cross-Lingual Transfer Learning (CLTL) in text classification tasks is to use a representation of words or documents extracted using a multilingual model; this representation should be independent of the language and capture high-level semantic and syntactic relations. Through transfer learning, it is possible to train a classifier using a dataset in one or more languages (source languages) and then transfer knowledge to different languages (target languages). This transfer learning approach is well-suited for low-resourced languages and for tasks requiring a lot of data. The performance obtained with CLTL aims to be as close as possible to training the entire system on language-specific resources.

There are different schemes for cross and multilingual document classification, which can be distinguished by the source and target languages, as well as the approach of selecting the best model. In a Zero-Shot Learning (ZSL) scheme, the source languages are different from the target languages, and the selection of the best model is performed using a development set from the source languages. In the Target Learning (TL) scheme, the source and target languages do not overlap, but the model selection is performed using the development set of target languages. In a Joint Learning (JL) scheme, the source and target languages are the same, and the selection method is applied using the development set of these languages.

\begin{figure}[htbp]
  \centering
  \includegraphics[width=\linewidth]{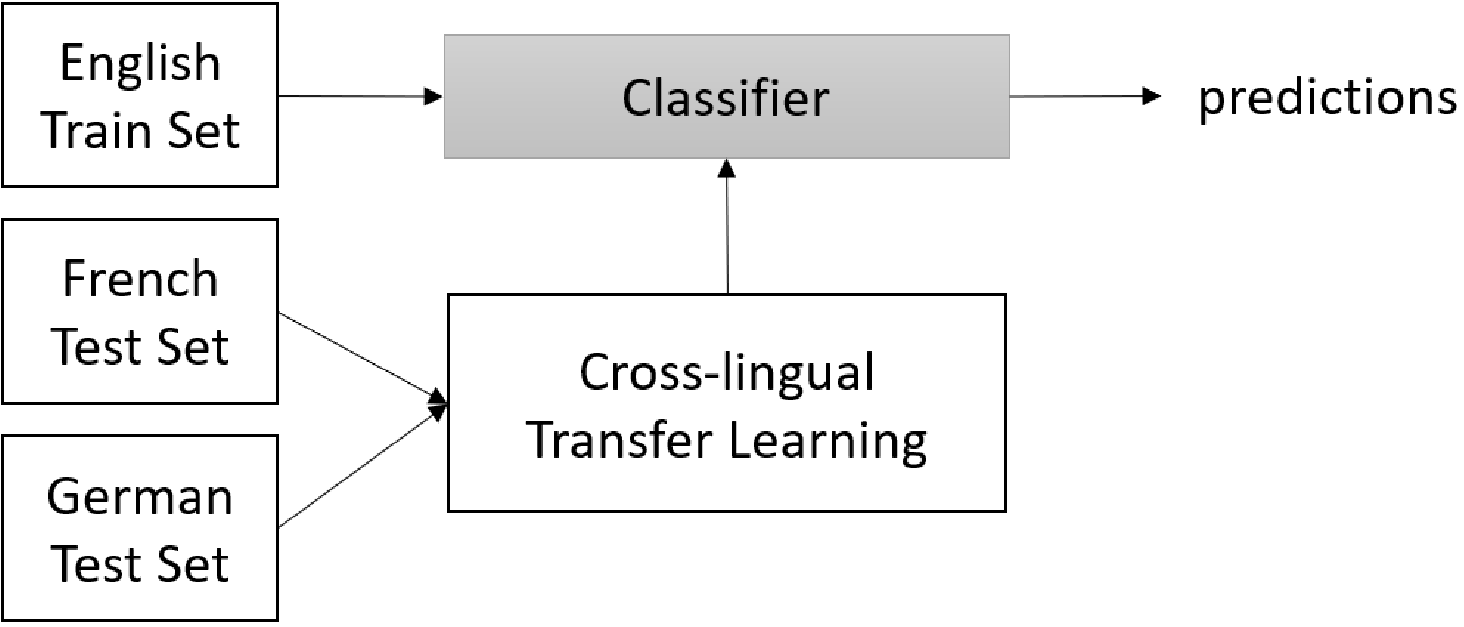}
  \caption{Zero-shot Cross-lingual transfer learning.}
\end{figure}


\section{Datasets}
\label{sec-datasets}
In this section, we introduce the multilingual EuroVoc thesaurus used to classify legal documents in both JRC-Acquis and EURLEX57K datasets. Then we explore the multilingual version of JRC-Acquis V3. We also describe how we extended EURLEX57K dataset by adding parallel documents available in French and German.

\subsection{EuroVoc Thesaurus}

The EuroVoc thesaurus is a multilingual thesaurus thematically covering many of the activities of the EU. It contains 20 domains, each domain contains a number of micro-thesauri. Descriptors in EuroVoc are classified under these micro-thesauri, and each descriptor belongs to one or more micro-thesauri. Relations between descriptors are represented using the SKOS ontology\footnote{https://www.w3.org/2004/02/skos}. Hierarchical relation between descriptors are specified with the SKOS \emph{broader} relation. 
Used instead relation identifies the relation between a descriptor and its replacement. 
The SKOS \emph{related} link is used to map a descriptor to its related descriptors; Used for relation maps each descriptor to its related labels. In total, there are 127 micro-thesaurus and 7221 Descriptors.


\subsection{JRC-Acquis multilingual}
\label{secjrc}

JRC-Acquis dataset is a smaller dataset with parallel documents in 20 languages; this dataset overlaps with EURLEX57K dataset and contains additional documents. It is labeled using descriptors from EuroVoc. We selected documents in English, French, and German for our experiments; we show statistics about this dataset in table~\ref{tab_stats_jrc}. We do not use unlabeled documents for classifier finetuning. Therefore, we do not assign them to any training split, and we use them only for language model finetuning.

\begin{table}[htbp]
\small
            \centering
            \caption{JRC-Acquis dataset in English (EN), French (Fr) and German (DE). Number of documents in train, development and test sets in addition to the number of documents with no split and the total number of documents.}
            \label{tab_stats_jrc}
            \begin{tabular}{|c|c|c|c|c|c|}
            \hline

Language & train & development & test & no split & total \\\hline

EN&16454&1960&1968&3163&23545\\\hline
FR&16434&1959&1967&3267&23627\\\hline
DE&16363&1957&1965&3256&23541\\\hline
\end{tabular}
        \end{table}
        
\subsection{EURLEX57K multilingual}
\label{seceurlex}
EUR-Lex documents are Legal documents from the European Union labeled using the set of EuroVoc thesaurus descriptors. 
We collected parallel documents in German and French to the documents in EURLEX57K dataset. We use the CELEX ID from the original EURLEX57K dataset to divide the data into train, development, and test sets. The documents from the parallel corpora are assigned the same splits as in the original monolingual EURLEX57K dataset. Therefore, our final dataset contains parallel texts in 3 languages. Statistics about this dataset are found in table~\ref{tab_stats_eur}.

\begin{table}[htbp]
\small
            \centering
            \caption{Multilingual EURLEX57K dataset in English (EN), French (Fr) and German (DE). Number of documents in train, development and test sets in addition to the number of documents with no split and the total number of documents.}
            \label{tab_stats_eur}
            \begin{tabular}{|c|c|c|c|c|c|}
            \hline

Language & train & development & test & no split & total \\\hline
EN&44428&5929&5921&24004&80282\\\hline
FR&44427&5929&5921&24452&80729\\\hline
DE&43749&5842&5820&23942&79353\\\hline
\end{tabular}
        \end{table}

We extended our dataset by including EUR-Lex documents that are not available in EURLEX57K. 
We use these additional documents only for Language Model finetuning stage (see section \ref{sub_sec_training_strategies}), so they do not have a training split, and we do not use them in classifier finetuning.

\section{Methods}
\label{section-methods}
In this section we describe the methods used in the ZSL and JT experiments presented in the results section, and multilingual training process.
Also, we discuss important related points such as language model finetuning, and gradual unfreezing.

\subsection{Multilingual Transformer Based Models}
\label{secmultimodels}

\textbf{BERT} it is a transformer-based architecture trained using masked language model (MLM) and next sentence prediction (NSP) objectives. In MLM, 15\% of the tokens are randomly masked, and the model tries to predict them from the context. BERT learns rich contextual representations of words and relations between them. 
BERT uses a special token [CLS] token for classification, which is added at the beginning of the text by the tokenizer. 
The token reflects the hidden representation of the last BERT layer and aggregates the sequence representation. 
BERT appeared in 2019, and since then was successfully applied in many natural language processing and understanding tasks. 
In this work, we utilize the multilingual version of BERT called M-BERT. \\
\textbf{DistillBERT} it is a distilled version of BERT; it achieves over 95\% of BERT's performance while having 40\% fewer parameters. In our experiments, we used DistillBERT to select the best training strategy for computationally expensive experiments, and then apply the strategy for M-BERT. We refer to the multilingual version of DistilBERT as M-DistilBERT.

\subsection{Multilingual Training}
To train our multilingual cross-lingual model, we finetune transformer-based models (see Section \ref{secmultimodels}) using multilingual documents from the legal domain (see Sections \ref{seceurlex} and \ref{secjrc}). 
The classifier is built upon the document representation produced by the M-BERT and M-DistilBERT models. We pass the representation of [CLS] token through a fully connected layer and then project the output to a vector with the size of the target classes count.\\
In finetuning the language model, we experimented with different numbers of epochs and different combinations of datasets; an ablation study is found in section \ref{ablation}.\\
The classifier is trained in the ZSL scheme using the English part of the dataset, we pick the model configuration with the best F1-score on the English test set, and evaluate it on the French and German datasets independently.\\

In JT scheme, the model is trained by including all the languages in the training and model picking process. 
We evaluate the selected model using the test sets in English, French, and German independently.
To evaluate the effect of having parallel languages in the training process, we compare the model trained in the ZSL scheme and the model trained in the JL scheme on the English test set; the results of this ablation study are given in Section \ref{ablation} 

\subsection{Training Strategies}
\label{sub_sec_training_strategies}

In line with Shaheen et al.~\cite{shaheen2020large}, we train multilingual classifiers using the training strategies described below. 
The first strategy is \emph{language model finetuning} of the transformer model before using it in classification. Finetuning is done on all training documents, and additionally on unlabeled documents available
in the EurLex database. This step aims at improving the model's representation of legal documents. 
Secondly, in \emph{gradual unfreezing}, we freeze all the model's layers with the exception of the last few layers. We start by training only those layers.
Later, the number of unfrozen layers is gradually increased during training. An ablation study about the effect of using such training strategies on multilingual models trained in ZSL and JT schemes is found in Section \ref{ablation}.
Both training strategies are proposed by Howard and Ruder \cite{howard2018universal}.

\subsection{Baseline}

Shaheen et al.~\cite{shaheen2020large} investigated the performance of various transformer-based models (including BERT, RoBERTa, and DistillBERT), in combination with training strategies such as language model finetuning and gradual unfreezing. The authors report their results on the English part of JRC-Acquis and EURLEX57K dataset. 
Here, we use these results as a baseline to compare our results on the English part of the datasets.\\
However, to the best of our knowledge, no baseline exists for the text classification using EurLex and JRC-Acquis for French and German, 
for which we provide a reference evaluation for both the JT and ZSL schemes. 

\subsection{Evaluation}

Following Shaheen et al. \cite{shaheen2020large}, we use F1-score as a decision support metric. This metric focuses on measuring how well the system helps to recommend correct labels, it aims at selecting relevant labels and avoiding irrelevant labels. Precision is the percentage of selected labels that are relevant to the document. Its focus is recommending mostly related labels. Recall is the percentage of relevant labels that the system selected. Its focus is not missing relevant labels. F1-score is the harmonic mean between precision and recall. These metrics have a major drawback, they are targeted at predicting relevant labels regardless their position in the list of predicted labels, and as a result this metric is not suitable for applications like a recommendation system. 

Shaheen et al. \cite{shaheen2020large} use additional retrieval measures for evaluation. R-Precision@K (RP@K), and Normalized Discounted Cumulative Gain (nDCG@K). These Rank-Aware metrics emphasis being good at finding and ranking labels, they rewards putting relevant labels high up in the list of recommendations and penalizes late recommendation of relevant labels.

\section{Results}
\label{sec_results}

This section reports the results of multilingual transformer-based models trained in the ZSL scheme (Section~\ref{sec_results_zsl}), the JT scheme (Section~\ref{sec_results_jt}) and the ablation studies (Section~\ref{sec_results}).  

\subsection{Zero-Shot Results}
\label{sec_results_zsl}

First, we evaluate multilingual transformer-based models (M-BERT and M-DistilBERT) trained in the ZSL scheme to classify French and German texts -- using only English texts as training data.
Table~\ref{tab_zero_shot_jrc} shows the results on JRC-Acquis dataset and, followed by Table~\ref{tab_zero_shot_eurlex} with the results for M-EURLEX57K. 
The French and German test sets are being evaluated separately.\\
In our experiments M-BERT consistently outperforms M-DistilBERT in the ZSL setting by a large margin across both datasets. 
Further, we observe better classification performance for the French datasets than for the respective German datasets, both for M-BERT and M-DistilBERT models.

\begin{table*}[htbp]
\caption{The results of multilingual models (M-BERT, M-DistilBERT) trained in ZSL scheme using the English part of JRC-Acquis on French (FR) and German (DE) parallel test sets.}
\begin{center}
\begin{tabular}{|c|c|c|c|c|c|c|c|c|}
\hline
        Language & Model & F1-score & RP@3 & RP@5 & nDCG@3 & nDCG@5\\\hline
        FR & M-DistilBERT & 0.504 & 0.628 & 0.56 & 0.66 & 0.604\\
        FR & M-BERT & \textbf{0.55} & \textbf{0.674} & \textbf{0.604} & \textbf{0.704} & \textbf{0.648}\\\hline\hline
        DE & M-DistilBERT & 0.473 & 0.583 & 0.527 & 0.613 & 0.566\\
        DE & M-BERT & \textbf{0.519} & \textbf{0.637} & \textbf{0.571} & \textbf{0.667} & \textbf{0.613}\\\hline
\end{tabular}
\label{tab_zero_shot_jrc}
\end{center}
\end{table*}

\begin{table*}[htbp]
    \centering
    \caption{The results of multilingual models (M-BERT, M-DistilBERT) trained in ZSL scheme using the English part of the multilingual EURLEX57K dataset on French (FR) and German (DE) test sets.}
    \label{tab_zero_shot_eurlex}
    \begin{tabular}{|c|c|c|c|c|c|c|c|c|}
        \hline
        Language & Model & F1-score & RP@3 & RP@5 & nDCG@3 & nDCG@5\\\hline
        FR & M-DistilBERT & 0.614 & 0.718 & 0.677 & 0.741 & 0.706\\
        FR & M-BERT & \textbf{0.67} & \textbf{0.771} & \textbf{0.726} & \textbf{0.795} & \textbf{0.757}\\\hline\hline
        DE & M-DistilBERT & 0.594 & 0.7 & 0.652 & 0.723 & 0.683\\
        DE & M-BERT & \textbf{0.648} & \textbf{0.751} & \textbf{0.7} & \textbf{0.776} & \textbf{0.733}\\\hline
    \end{tabular}
\end{table*}

\subsection{Joint Training Results}
\label{sec_results_jt}

We continue with the evaluation of multilingual transformer-based Models (M-BERT and M-DistilBERT) trained in the JT scheme for English, French and German languages. 
The results of monolingual models (BERT, RoBERTa, and DistilBERT), as reported in Shaheen et al.~\cite{shaheen2020large}, serve as a baseline on the English test set. 

\textbf{JRC Acquis:} Table~\ref{tab_jrc_results} presents an overview of the results on JRC-Acquis. We observe that transformer-based models, trained using JRC Acquis in the JT scheme, fail to reach the performance of monolingual models on the English test set. In this manner, multilingual models achieve about 96.83-98.39\% of the performance achieved by monolingual baseline models. Interestingly, both M-DistilBERT and M-BERT perform similarly according to all metrics, with slightly better performance for M-BERT on F1-score and slightly better performance for M-DistilBERT on the rest of the metrics (RP@3, RP@5, nDCG@3, nDCG@5).

\begin{table*}[t]
            \centering
             \caption{M-BERT and M-DistilBERT results trained in the JT scheme for the JRC Acquis dataset in English (EN), French (FR) and German(DE), plus baseline results of monolingual models (BERT, DistilBERT, RoBERTa) on the English test set.}
           \label{tab_jrc_results}
            \begin{tabular}{|c|c|c|c|c|c|c|}
            \hline

Language & Model & F1-score & RP@3 & RP@5 & nDCG@3 & nDCG@5\\\hline

FR & M-DistilBERT & 0.637 & \textbf{0.766} & 0.692 & \textbf{0.79} & 0.732\\
FR & M-BERT & \textbf{0.642} & 0.763 & \textbf{0.696} & 0.785 & \textbf{0.733}\\\hline\hline

DE & M-DistilBERT & 0.634 & \textbf{0.762} & 0.691 & \textbf{0.787} & \textbf{0.731}\\
DE & M-BERT & \textbf{0.641} & 0.759 & \textbf{0.693} & 0.781 & 0.729\\\hline\hline

EN & M-DistilBERT & 0.638 & 0.768 & 0.697 & 0.794 & 0.737\\
EN & M-BERT & 0.644 & 0.763 & 0.695 & 0.785 & 0.733\\\hline

EN & DistilBERT & 0.652  & 0.78 & 0.711 & 0.805 & 0.75\\
EN & BERT & \textbf{0.661} & 0.784 & 0.715 & 0.803 & 0.750\\
EN & RoBERTa & 0.659 & \textbf{0.788}  & \textbf{0.716} & \textbf{0.807} & \textbf{0.753}\\\hline

\end{tabular}
\end{table*}

\textbf{EURLEX57K:} In contrast to JRC-Acquis, for the M-EURLEX57K (see Table~\ref{tab_eur_results})
M-BERT achieves similar or slightly better results on all metrics than RoBERTa (the best baseline model),
when comparing multilingual models to the monolingual baseline.
Also, M-BERT provides an improvement of 1\% over monolingual (English) BERT on all metrics. Although monolingual DistilBERT achieves slightly better results than M-DistilBERT, results are also identical. 

\begin{table*}[htbp]
\small
            \centering
            \caption{M-BERT and M-DistilBERT results trained in the JT scheme for the EURLEX57K dataset in English (EN), French (FR) and German(DE), plus baseline results of monolingual models (BERT, DistilBERT, RoBERTa) on the English test set.}
            \label{tab_eur_results}
            \begin{tabular}{|c|c|c|c|c|c|c|}
            \hline

Language & Model & F1-score & RP@3 & RP@5 & nDCG@3 & nDCG@5\\\hline

FR & M-DistilBERT & 0.754 & 0.846 & 0.803 & 0.864 & 0.829\\
FR & M-BERT & \textbf{0.761} & \textbf{0.851} & \textbf{0.811} & \textbf{0.867} & \textbf{0.833}\\\hline\hline

DE & M-DistilBERT & 0.751 & 0.843 & 0.801 & 0.862 & 0.827\\
DE & M-BERT & \textbf{0.759} & \textbf{0.847} & \textbf{0.807} & \textbf{0.864} & \textbf{0.831}\\\hline\hline

EN & M-DistilBERT & 0.753 & 0.847 & 0.803 & 0.865 & 0.829\\
EN & M-BERT & \textbf{0.761} & \textbf{0.85} & \textbf{0.812} & \textbf{0.867} & \textbf{0.836}\\\hline

EN & DistilBERT & 0.754 & 0.848 & 0.807 & 0.866 & 0.833\\
EN & BERT & 0.751 & 0.843 & 0.805 & 0.859 & 0.828\\
EN & RoBERTa & 0.758 & \textbf{0.85} & \textbf{0.812} & 0.866 & 0.835\\\hline
\end{tabular}
\end{table*}

\subsection{Ablation Studies}
\label{ablation}

In this set of experiments, we study the contributions of different training components and training strategies on the ZSL model -- by excluding some of those components individually or reducing the number of training epochs. We focus on three components: 
(i) the use of gradual unfreezing or not, (ii) the number of unfrozen layers, (iii) and the number of language model finetuning epochs. 
In all those experiments, we train the models using the English training data of JRC Acquis, and we test using French and German test sets.\\

Table~\ref{tab_abl_nogduf} provides a comparison of the evaluation metrics with or without gradual unfreezing. For both French and German, we can see consistent a improvement of results when using gradual unfreezing. The relative improvement for French is in the range 38-45\%, and for German is in the range 58-70\%. In conclusion, gradual unfreezing is a crucial component for good classification performance of a model trained in the ZSL scheme.\\
Next, we examine the effect of freezing the network layers at the start of training, and gradually unfreezing some the of the layers during training (Table~\ref{tab_abl_gduf}).

\begin{table*}[htbp]
\small
            \centering
        \caption{Ablation Study: ZSL M-DistilBERT performance on JRC-Acquis depending on the number the unfrozen layers. Again, we train on the English training set, and test on French and German.}
            \label{tab_abl_gduf}
            \begin{tabular}{|c|c|c|c|c|c|c|c|c|}
            \hline

Language & Unfrozen Layers & F1-score & RP@3 & RP@5 & nDCG@3 & nDCG@5\\\hline

FR & Last 2 layers & 0.434 & 0.543 & 0.486 & 0.574 & 0.527\\
FR & Last 3 layers & 0.442 & 0.547 & 0.493 & 0.58 & 0.533\\
FR & Last 4 layers & 0.439 & 0.549 & 0.491 & 0.579 & 0.532\\
FR & Last 5 layers & \textbf{0.455} & \textbf{0.567} & \textbf{0.505} & \textbf{0.597} & \textbf{0.547}\\
FR & All 6 layers & 0.451 & 0.563 & 0.5 & 0.593 & 0.542\\
FR & All 6 layers + EMB & \textbf{0.455} & 0.566 & 0.504 & 0.596 & 0.546\\\hline

DE & Last 2 layers & 0.388 & 0.471 & 0.429 & 0.501 & 0.463\\
DE & Last 3 layers & 0.393 & 0.484 & 0.434 & 0.509 & 0.468\\
DE & Last 4 layers & 0.381 & 0.466 & 0.418 & 0.495 & 0.454\\
DE & Last 5 layers & \textbf{0.395} & \textbf{0.488} & \textbf{0.442} & \textbf{0.516} & \textbf{0.477}\\
DE & All 6 layers & 0.384 & 0.468 & 0.42 & 0.497 & 0.456\\
DE & All 6 layers + EMB & 0.391 & 0.474 & 0.428 & 0.504 & 0.464\\\hline
\end{tabular}
        \end{table*}

Gradually unfreezing the last five layers while keeping the first and embedding (EMB) layers frozen achieves the best performance on French and German test sets. 
Unfreezing all layers (including the embedding layer) obtains very close results to the best results on the French test set, while the difference on the German test set is a bit larger.\\
In Table~\ref{tab_abl_lmft}, we test the effect of the number of language model finetuning epochs. 
On the French test set, one cycle of language model finetuning leads to 18.6-20.48\% of relative gain compared to no LM finetuning at all. Increasing the number of epochs to 5 and 10 increases the relative gain to 29.6-32.53\% and 32.0-34.94\% correspondingly. The difference is much bigger on the German test set, compared to no LM finetuning the relative gain is 42.82-49.47\%, 70.69-81.49\%, 76.15-87.54\% for 1, 5, 10 epochs of LM finetuning. 

\begin{table*}[htbp]
\small
            \centering
        \caption{Ablation Study: ZSL M-DistilBERT performance on JRC-Acquis depending on the number of language model finetuning cycles (LMFT-cycles) -- with 6 layers and unfrozen and training on the English training set.}
            \label{tab_abl_lmft}
            \begin{tabular}{|c|c|c|c|c|c|c|c|c|}
            \hline

Language & \#LMFT-cycles & F1-score & RP@3 & RP@5 & nDCG@3 & nDCG@5\\\hline

FR & 0 & 0.379 & 0.47 & 0.415 & 0.5 & 0.454\\
FR & 1 & 0.451 & 0.563 & 0.5 & 0.593 & 0.542\\
FR & 5 & 0.498 & 0.615 & 0.55 & 0.648 & 0.595\\
FR & 10 & \textbf{0.504} & \textbf{0.628} & \textbf{0.56} & \textbf{0.66} & \textbf{0.604}\\\hline

DE & 0 & 0.267 & 0.32 & 0.281 & 0.348 & 0.313\\
DE & 1 & 0.384 & 0.468 & 0.42 & 0.497 & 0.456\\
DE & 5 & 0.459 & 0.563 & 0.51 & 0.594 & 0.549\\
DE & 10 & \textbf{0.473} & \textbf{0.583} & \textbf{0.527} & \textbf{0.613} & \textbf{0.566}\\\hline
\end{tabular}
        \end{table*}
        
\begin{table*}[t]
\small
            \centering
    \caption{Ablation Study: ZSL M-DistilBERT performance on JRC-Acquis regarding the use of gradual unfreezing (GDUF). We unfreeze 6 layers and train on the English training set.}
            \label{tab_abl_nogduf}
            \begin{tabular}{|c|c|c|c|c|c|c|c|c|}
            \hline

Language & GDUF & F1-score & RP@3 & RP@5 & nDCG@3 & nDCG@5\\\hline

FR & False & 0.327 & 0.385 & 0.351 & 0.406 & 0.377\\
FR & True & \textbf{0.451} & \textbf{0.563} & \textbf{0.5} & \textbf{0.593} & \textbf{0.542}\\\hline

DE & False & 0.243 & 0.274 & 0.248 & 0.291 & 0.267\\
DE & True & \textbf{0.384} & \textbf{0.468} & \textbf{0.42} & \textbf{0.497} & \textbf{0.456}\\\hline
\end{tabular}
        \end{table*}
    
\section{Discussion}
\label{sec-discussion}

We included much of the detailed discussion in
the results section (Section \ref{sec_results}), so here we will summarize and extend on some of the key findings.\\
Comparing the results of the ZSL scheme (Tables ~\ref{tab_zero_shot_jrc} and \ref{tab_zero_shot_eurlex}) to the JT scheme (Tables ~\ref{tab_jrc_results} and \ref{tab_eur_results}) on French and German test sets, the experiments show that M-BERT trained in ZSL scheme reaches about 86\% of the performance of a model trained in the JT scheme. In the same way, M-DistilBERT in ZSL settings achieves about 79\% of the performance of the JT scheme.

Additionally, the multilingual models (M-BERT, M-DistilBERT) trained in the JT scheme on English, French and German provide similar performance on their respective test sets (see tables \ref{tab_jrc_results} and \ref{tab_eur_results}).
However, when using the ZSL scheme, there is a discrepancy of between French and German results, indicating that the multilingual models can more easily transfer from the English to the French representations (Tables \ref{tab_zero_shot_jrc} and \ref{tab_zero_shot_eurlex}). 

\section{Conclusion}
\label{sec-conclusion}

In this work, we evaluate cross-lingual transfer learning for LMTC task, based on JRC-Acquis dataset and an extended version of EURLEX57K dataset. We started a baseline for LMTC based on these two multilingual datasets which contain parallel documents in English, French and German. We also compared between two CLTL settings for this task:  zero-shot setting and joint training setting.

The main contributions of this work are: (i) the experiments with multilingual-BERT and multilingual-DistilBERT with gradual unfreezing and language model finetuning (ii) providing a new standardized multilingual dataset for further investigation, (iii) ablation studies to measure the impact and benefits of various training strategies on Zero-shot and Joint-Training transfer learning.

There are multiple angles for future work, including potentially deriving higher performance by using hand-picked learning rates and other hyperparameters for each model individually. Moreover, experiments with language adversarial training and various data augmentation techniques are candidates to improve classification performance. 

\bibliographystyle{IEEEtran}
\bibliography{conference_101719}

\end{document}